\documentclass[10pt,journal,twocolumn,twoside]{IEEEtran} 
\usepackage{graphicx}
\usepackage{epstopdf}
\usepackage{float}
\usepackage{algorithmic}
\usepackage{array}
\usepackage{amsmath}
\usepackage{amssymb}
\usepackage{mdwmath}
\usepackage{booktabs}
\usepackage{eqparbox}
\usepackage{stfloats}
\usepackage{hyperref}
\usepackage{cleveref}
\hypersetup{hypertex=true,
            colorlinks=true,
            linkcolor=blue,
            anchorcolor=blue,
            citecolor=blue}
\usepackage{cases} 
\usepackage{subfigure}
\usepackage{upgreek}
\usepackage{multirow}
\usepackage{makecell}
\usepackage{amsmath}
\usepackage[boxed,ruled,commentsnumbered]{algorithm2e}
\usepackage{url}
\usepackage[table,xcdraw]{xcolor}
\usepackage{colortbl}
\usepackage{cite}
\ifCLASSOPTIONcompsoc
\usepackage[caption=false,font=normalsize,labelfont=sf,textfont=sf]{subfig}
\else
\fi
\allowdisplaybreaks[4]

\makeatletter

\renewcommand*{\@opargbegintheorem}[3]{\trivlist
      \item[\hskip \labelsep{\bfseries #1\ #2}] \textbf{(#3):}\ }
\makeatother

\begin{document}
\title
{Empower Low-Altitude Economy: A Reliability-Aware Dynamic Weighting Allocation for Multi-modal UAV Beam Prediction}

\author{Haojin Li, Anbang Zhang, Chen Sun,~\IEEEmembership{Senior Member, IEEE}, Chenyuan Feng,~\IEEEmembership{Senior Member,~IEEE}, Kaiqian Qu, \\ Tony Q. S. Quek,~\IEEEmembership{Fellow, IEEE} and Haijun Zhang,~\IEEEmembership{Fellow, IEEE} 
\thanks{
(\emph{*Corresponding author: Anbang Zhang})

Haojin Li and Haijun Zhang are with University of Science and Technology Beijing, China (email: Haojin.li@sony.com, haijunzhang@ieee.org). 

Haojin Li and Chen Sun are with Sony China Research Laboratory, China (email:  chen.sun@sony.com).

Anbang Zhang is with School of Control Science and Engineering, Shandong University, China (e-mail: zab\_0613@163.com). 

Kaiqian Qu is with Southeast University, Nanjing 210096, China
(e-mail:  qukaiqian2021@163.com).

Chenyuan Feng is with the College of Computer Science, University of Exeter, U.K. (email: c.feng@exeter.ac.uk).

T.~Q.~S.~Quek is with the Information Systems Technology and Design Pillar, Singapore University of Technology and Design, Singapore 487372 (e-mail: tonyquek@sutd.edu.sg).
}
}
\maketitle


\begin{abstract} 
The low-altitude economy (LAE) is rapidly expanding driven by urban air mobility, logistics drones, and aerial sensing, while fast and accurate beam prediction in uncrewed aerial vehicles (UAVs) communications is crucial for achieving reliable connectivity. Current research is shifting from single-signal to multi-modal collaborative approaches. However, existing multi-modal methods mostly employ fixed or empirical weights, assuming equal reliability across modalities at any given moment. Indeed, the importance of different modalities fluctuates dramatically with UAV motion scenarios, and static weighting amplifies the negative impact of degraded modalities. Furthermore, modal mismatch and weak alignment further undermine cross-scenario generalization.
To this end, we propose a reliability-aware dynamic weighting scheme applied to a semantic-aware multi-modal beam prediction framework, named $\text{SaM²B}$. 
Specifically, $\text{SaM²B}$ leverages lightweight cues such as environmental visual, flight posture, and geospatial data to adaptively allocate contributions across modalities at different time points through reliability-aware dynamic weight updates.
Moreover, by utilizing cross-modal contrastive learning, we align the “multi-source representation beam semantics” associated with specific beam information to a shared semantic space, thereby enhancing discriminative power and robustness under modal noise and distribution shifts. Experiments on real-world low-altitude UAV datasets show that $\text{SaM²B}$ achieves more satisfactory results than baseline methods.
\end{abstract}


\section{Introduction} 

\IEEEPARstart{W}{ith} the requirement of sixth-generation (6G) mobile communications, the world is entering an unprecedented era of massive data interaction \cite{10443316}. However, constrained by spectrum resources and insufficient deployment density of ground base stations (BSs), traditional cellular networks still face bottlenecks in achieving ubiquitous coverage and large-scale concurrent communications. LAE, as an emerging application paradigm, integrates low-altitude aviation activities with advanced air mobility aircraft, providing seamless, three-dimensional services for scenarios such as airborne networks \cite{11017717} and smart transportation \cite{11088255}. 

To support such applications, low-altitude UAV networks must possess reliable high-speed communication and real-time sensing capabilities. Notably, ITU-R has approved the IMT-2030 (6G) draft recommendation, identifying integrated sensing and communication (ISAC) as one of the 6G key application scenarios \cite{10977743}, aiming to widely deploy wireless sensing architectures in future networks. This means that future low-altitude communication networks will not only provide high-speed data links for UAVs but also enable real-time sensing \cite{10950390} of the surrounding airspace environment, assisting in obstacle avoidance, navigation, and mission execution. 

However, the low-altitude high-dynamic environments pose significant challenges, despite the widespread deployment of mmWave links offers the advantage of ultra-wide bandwidth, it still suffers from severe path loss and rapid channel changes caused by the high-speed maneuvering of UAVs. Also, complex urban terrain further exacerbates link instability. To maintain a robust connection, the system must complete beam prediction and switching within milliseconds. Traditional beam management methods based on periodic channel detection \cite{8025577} are too costly and have unacceptable latency in LAE. Single Location-based beam selection is highly dependent on high-precision positioning. Recently, deep learning (DL) methods have been introduced into beam prediction \cite{9149272}, attempting to directly map sensor data to beam indices. However, most existing solutions rely on a single modality \cite{9512383}, resulting in a significant performance decline when modality degradation occurs. More critically, even within multi-modal systems, existing methods often employ fixed fusion weights, assuming all modalities are equally reliable at any given moment. In practice, modal reliability varies significantly across scenarios. Without explicitly modeling modal importance, fusion struggles to adapt. Consequently, there is an urgent need for a more reliable multi-modal adaptation strategy capable of measuring the importance of different modalities.

Motivated by this, we propose a reliability-aware, semantic multi-modal beam prediction framework, $\text{SaM²B}$. Specifically, $\text{SaM²B}$ incorporates lightweight cues such as environmental visual information, geospatial cues, and flight attitude features. It then employs attention mechanisms to infer the reliability of each modality, enabling adaptive weight allocation at every time step. Moreover, by bringing paired samples closer and unpaired samples farther apart, it generates modality-invariant and discriminative representations. 
Comprehensive experiments on real-world low-altitude UAV datasets demonstrate that $\text{SaM²B}$ consistently outperforms existing baseline methods, while maintaining computational efficiency suitable for airborne deployment.

\begin{figure*}[htbp]
\centering
\includegraphics[width=1\linewidth]{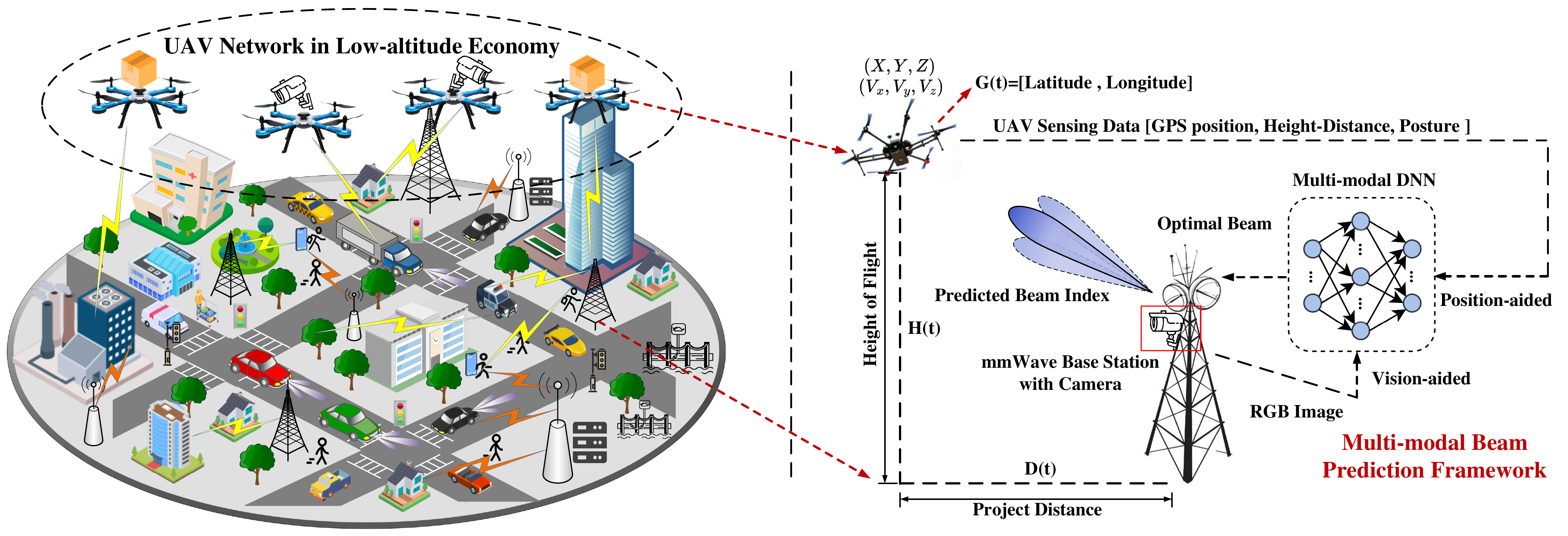}
\caption{UAV network in low-altitude economy and the proposed multi-modal beam prediction framework.}
\label{fig1}
\end{figure*}

\section{System Model and Problem Formulation} 


\subsection{System Model}

As shown in Fig. \ref{fig1}, we consider a UAV mmWave communication system operating in the LAE scenario. This communication system employs orthogonal frequency-division multiplexing (OFDM) transmission with $K$ subcarriers and a cyclic prefix of length $D$. In the downlink, the BS applies beamforming to enhance the received signal quality at the UAV. To enable directional transmission, the BS employs a pre-defined beamforming codebook $\mathcal{F}=\left\{\mathbf{f}_{q}\right\}_{q=1}^{Q}$, where $\mathbf{f}_{q} \in \mathbb{C}^{M \times 1}$ denotes the $q$-th beamforming vector and $Q$ is the total number of candidate beams. Let $\mathbf{h}_{k}[t] \in \mathbb{C}^{M \times 1}$ denote the downlink channel vector between the BS and the UAV at the $k$-th subcarrier and time instant $t$. 

Accordingly, the received signal at the UAV on the $k$-th subcarrier is given by: $y_k[t] = \mathbf{h}_k^H[t]\mathbf{f}_{q[t]}s + n_k[t]$,
where $s\in \mathbb{C}$ is the transmitted complex symbol with power constraint $\mathbb{E}[|s|^2] = P$, $P$ denotes the average symbol power. And $\mathbf{h}_k^H$ is the mmWave channel vector, and $n[t] \sim \mathcal{C} \mathcal{N}\left(0, \sigma_{n}^{2}\right)$ represents complex additive Gaussian noise. The selection strategy of the optimal beam index $q^*[t]$ that maximizes the average receive signal-to-noise ratio (SNR) over all $K$ subcarriers, i.e.,
\begin{equation}\label{eq2}
q^{*}[t]=\arg \max _{q \in Q} \frac{1}{K} \sum_{k=1}^{K} \frac{P}{\sigma^{2}}\left|\mathbf{h}_{k}^{H}[t] \mathbf{f}_{{q[t]}}\right|^{2} .
\end{equation}
where $q^{*}[t]$ denotes the index of the optimal beamforming vector at time $t$.
Here, ${P}/{\sigma^{2}}$ corresponds to the transmit SNR. Thus, the selected beamforming vector is $\mathbf{f}^{*}[t]=\mathbf{f}_{q^{*}[t]}$. This selection criterion ensures the highest possible received SNR under the given codebook constraints.

\subsection{Problem Description}
At time $t$, we can obtain observations and representations of $S$ modalities ($S \ge 1$). For example, environmental vision, geographic location, and flight attitude. Conventional approaches typically cascade multi-modal features or perform linear fusion with fixed weights:
\begin{equation}
\mathbf{z}(t)=\sum_{s=1}^{S} \bar{w}_{s} \mathbf{F}_{s} \mathbf{X}_{s}(t).
\end{equation}
where, $\bar{w}_{s}=1$ is independent of time or context. This implies the assumption that all modalities are equally reliable at any time $t$. And $\mathbf{F}_{s}$ is a mapping transformation in neural networks. $\mathbf{X}_{s}(t)$ is the encoder output for one modality.
\begin{itemize}
\item \emph{The inherent fragility of single modality:} Models relying solely on a single sensor source (such as GPS alone or IMU alone) are highly prone to failure in real-world environments. Changes in single-modality distribution can cause learned maps to rapidly degrade, necessitating supplementary data from other modalities.
\item \emph{A key shortcoming in traditional multi-modal fusion:} To enhance robustness, many approaches incorporate multi-modal data, but they often employ static fusion through concatenation with fixed weights. This overlooks the varying confidence levels across different encoder outputs, and averaging their outputs with equal weights leads to suboptimal decisions.
\end{itemize}

In contrast, our reliability-focused multi-modal approach does not rely on stable channel statistics. Instead, it dynamically evaluates the credibility of each modality to transform heterogeneous sensor cues into a unified and semantically consistent signal. This transforms beam prediction from passive channel fitting into active environment-driven inference, enabling rapid and accurate UAV beam prediction.

\section{$\text{SaM²B}$: Reliability-Aware Dynamic Weighting for Robust Multi-Modal Beam Prediction} 

\begin{figure}[t]
\centering
\includegraphics[width=1\linewidth]{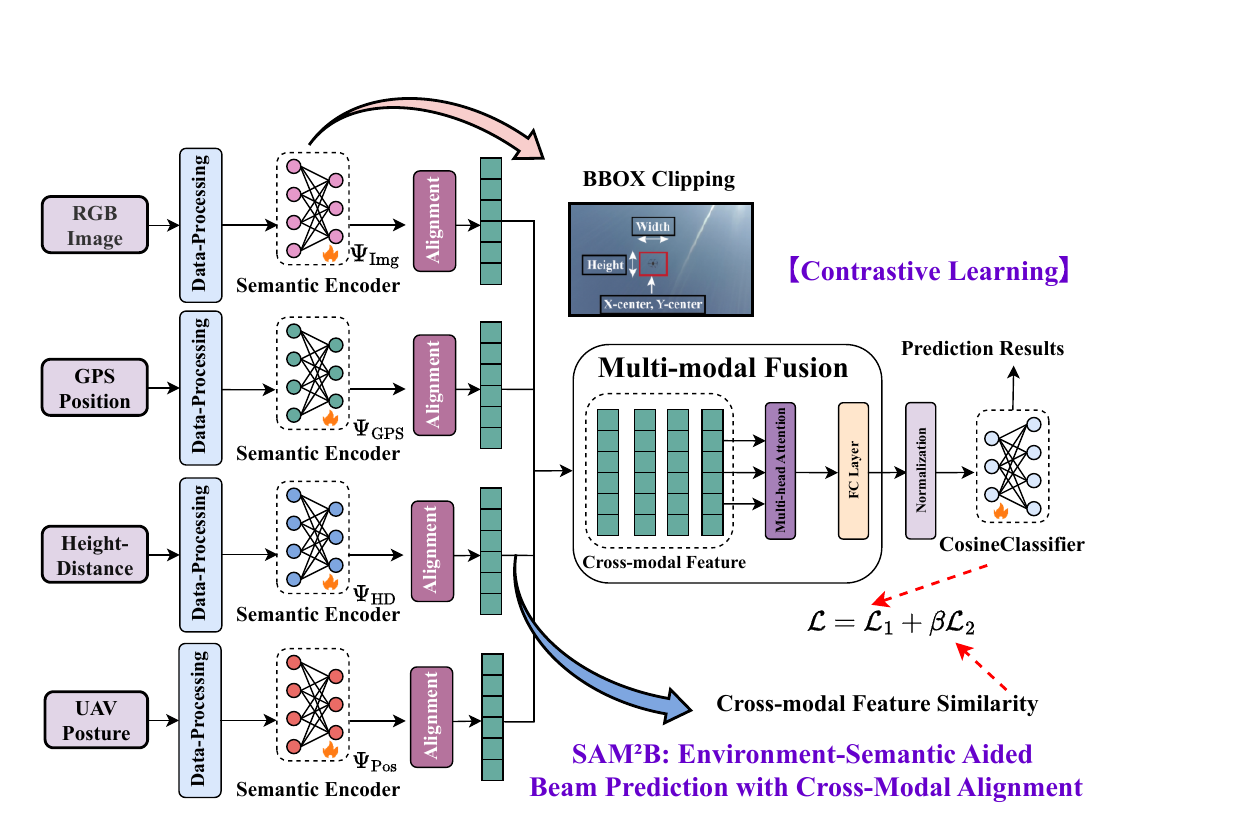}
\caption{$\text{SaM²B}$ performs semantic-aware multi-modal beam prediction by dynamically reweighting modalities under reliability cues.}
\label{fig2}
\end{figure}


    
        
  
  

  
  

\subsection{Multi-modal Feature Encoding}
Within the above system model, we propose a novel beam prediction framework that leverages sensory data collected at either the BS or the UAV, as shown in Fig. 2. 

Specifically, we set the BS to obtain the following sensory data: \emph{RGB Image:} The BS camera captures the visual scene as an RGB image, defined as $\mathbf{X}[t] \in \mathbb{R}^{W \times H \times C}$, where $W$, $H$, and $C$ represent the image width, height, and RGB spectral channels, respectively. We choose to perform BBOX clipping on the original RGB images. \emph{GPS Position:} The UAV collects its latitude and longitude information via GPS and feeds it back to the BS, defined as $\mathbf{G}[t]=[\text{Latitude, Longitude}] \in \mathbb{R}^{2}$ as the UAV position vector at time $t$. \emph{Height and Distance:} The UAV relative altitude and horizontal displacement with respect to the BS are jointly expressed as $\mathbf{HD}[t]=[\mathbf{H}[t], \mathbf{D}[t]]\in \mathbb{R}^{2}$. \emph{UAV Posture:} The UAV instantaneous orientation (e.g., roll, pitch, yaw angles) is defined as $\mathbf{V}[t] \in \mathbb{R}^{3}$, capturing its real-time posture dynamics.

Let us denote the sensory data at time $t$ as a multi-modal set $A[t]=\{\mathbf{X}[t], \mathbf{G}[t], \mathbf{HD}[t], \mathbf{V}[t]\}$. Specifically, we formalize the feature extraction process for each sensing modality. A detailed encoding description is given below:

\textbf{\emph{1) Visual Semantics With BBOX Clipping:}} Let $\mathbf{b}[t]=\left[x_{c}, y_{c}, w, h\right]$ denote the normalized BBOX produced by the link detector at time $t$. Based on this BBOX, we crop a ROI  from the RGB frame $\mathbf{X}[t]$ and resize it to $224\times 224$ via RoIAlign:
\begin{equation}
\mathbf{X}^{\text{roi}}[t]=\text{RoIAlign}(\mathbf{X}[t],\mathbf{b}[t]) \in \mathbb{R}^{224\times224\times3}.
\end{equation}

After ImageNet normalization, the ROI is passed through a ResNet-18 backbone $\Psi_{\text{img}}$ (with classification head removed), followed by a lightweight MLP head $\Gamma_{\text{img}}$, yielding a $Q$-dimensional embedding:
\begin{equation}
\tilde{\mathbf{X}}_{\text{Img}}[t]=\Gamma_{\text{Img}}\left(\Psi_{\text{Img}}\left(\mathbf{X}^{\text{roi}}[t]\right)\right)\in \mathbb{R}^Q.
\end{equation}

\emph{Remark.} If $\mathbf{b}[t]$ is unavailable, the framework defaults to global average pooling over the full frame, ensuring robustness by trading precision for coverage.

\textbf{\emph{2) GPS Position Information:}} During flight, UAVs continuously acquire the geographic coordinates $\mathbf{G}[t]$. After normalization, a lightweight multilayer perceptron (MLP) $\Psi_{\text{GPS}}$ maps the input into a $Q$-dimensional shared representation space to extract semantic position features:
\begin{equation}
\tilde{\mathbf{X}}_{\mathrm{GPS}}[t]=\Psi_{\mathrm{GPS}}(\mathbf{G}[t]) \in \mathbb{R}^{Q}.
\end{equation}

This transformation preserves geospatial information while reducing the sparsity of raw coordinates in the high-dimensional embedding space.

\textbf{\emph{3) UAV Height and Horizontal Distance:}} The UAV relative altitude $\mathbf{H}[t]$ and horizontal displacement $\mathbf{D}[t]$ with respect to the base station jointly form a two-dimensional feature vector $\mathbf{HD}[t]$. After min–max normalization, an MLP $\Psi_{\text{HD}}$ projects it into the shared semantic space:
\begin{equation}
\tilde{\mathbf{X}}_{\mathrm{HD}}[t]=\Psi_{\mathrm{HD}}(\mathbf{H D}[t]) \in \mathbb{R}^{Q}.
\end{equation}

The resulting embedding encodes the UAV–BS spatial geometry, serving as a crucial constraint for beam prediction.

\textbf{\emph{4) UAV Posture:}} UAV posture is obtained from the onboard inertial measurement unit (IMU), which provides real-time orientation parameters including roll, pitch, and yaw, denoted as $\mathbf{V}[t] = [\text{roll}, \text{pitch}, \text{yaw}] \in \mathbb{R}^3$. After normalization, an MLP $\Psi_{\text{att}}$ embeds the posture vector into the semantic space:
\begin{equation}
\tilde{\mathbf{X}}_{\mathrm{Pos}}[t]=\Psi_{\mathrm{Pos}}(\mathbf{V}[t]) \in \mathbb{R}^{Q}.
\end{equation}

This representation captures UAV orientation dynamics, enabling the framework to account for mobility-induced fluctuations in channel characteristics.

\subsection{Contrastive Learning–Driven Multi-modal Consistency}

To avoid the degradation of certain modalities in multi-modal fusion from fixed weights, we introduce credibility-aware dynamic weights. At each time step, fusion coefficients are adaptively assigned based on modal quality, thereby emphasizing modalities with higher informational value. 

\emph{Cross-modality Alignment:} $\text{SaM²B}$ utilizes cosine similarity constraints to align the different modality representations, inspired by CLIP \cite{pmlr-v139-radford21a}.

Specifically, given a semantic feature vector $\tilde{\mathbf{X}}_\gamma[t]$ extracted from modality $\gamma \in A$, we first perform $\ell_2$-normalization:
\begin{equation}
\bar{\mathbf{X}}_\gamma[t] = \frac{\tilde{\mathbf{X}}_\gamma[t]}{\|\tilde{\mathbf{X}}_\gamma[t]\|_2}, \quad \forall \gamma \in S,
\end{equation}
which projects all features onto the unit hypersphere. This normalization stabilizes the training process by constraining the feature space and allows cosine similarity to be directly applied as the alignment metric.

To capture cross-modal consistency, we compute the pairwise similarity between modalities $\gamma_1$ and $\gamma_2$ at time $t$ as
\begin{equation}
S_{\gamma_1,\gamma_2}[t] = \bar{\mathbf{X}}_{\gamma_1}[t]^\top \bar{\mathbf{X}}_{\gamma_2}[t], \quad \forall \gamma_1, \gamma_2 \in A,
\end{equation}
where $S[t] \in [-1,1]^{|A|\times|A|}$ forms a symmetric similarity matrix. Maximizing these pairwise similarities encourages modality-invariant representations, effectively disentangling semantic features from modality-specific noise. 

By encouraging cross-modal similarity, this geometric constraint in $\text{SaM²B}$ implicitly disentangles modality-invariant semantics from modality-specific noise, enhancing robustness and generalizability under the distribution shifts. 

\emph{Cross-Modality Attention Mechanism:} We stack the normalized features $\{\bar{\mathbf{X}}_{\gamma}[t]\}$ of each modality into a matrix $\mathbf{F}[t]= [\bar{\mathbf{X}}_{\text{Img}}^{\top}[t], \bar{\mathbf{X}}_{\text{GPS}}^{\top}[t], \bar{\mathbf{X}}_{\text{HD}}^{\top}[t], \bar{\mathbf{X}}_{\text{Pos}}^{\top}[t]]^{\top} \in \mathbb{R}^{4 \times Q} $ and model cross-modal dependencies using the transformer-based multi-head self-attention mechanism. Under this operation, queries, keys, and values are derived from this feature matrix $\mathbf{F}[t]$, i.e., $\mathbf{F}_Q(t)=\mathbf{F}_K(t)=\mathbf{F}_V(t)=\mathbf{F}^{\top}[t]$, to achieve the capture of dynamic correlation patterns across modalities. The correlations extracted by each attention head are spliced and projected back to the original space as
\begin{equation}
\mathbf{F}_{\text{Att}}[t]=\operatorname{MultiHead}\left(\mathbf{F}^{\top}[t], \mathbf{F}^{\top}[t], \mathbf{F}^{\top}[t]\right) \in \mathbb{R}^{|A| \times Q}.
\end{equation}

\emph{Cross-Modality Attention:}
Due to the different capabilities of each modality, we allow dynamic weighting and semantic sensing fusion of specific modality features. Let $\mathbf{v}_{s}[t]=\mathbf{F}_{\mathrm{Att}}[t]_{(s,:)} \in \mathbb{R}^{Q}$. We first obtain an attention score from data,
\begin{equation}
f_{s}[t]=\operatorname{MLP}_{s}\left(\mathbf{v}_{s}[t]\right),
\end{equation}
and a reliability score from lightweight quality cues,
\begin{equation}
c_{s}[t]=\operatorname{MLP}_{c}\left(\mathbf{r}_{s}[t]\right),
\end{equation}
where $\mathbf{r}_{s}[t]$ collects modality-specific quality features. To adaptively emphasize reliable modalities, we fuse two scores:
\begin{equation}
\tilde{f}_{s}[t]=\alpha f_{s}[t]+(1-\alpha) \phi\left(c_{s}[t]\right),
\end{equation}
and
\begin{equation}
w_{s}[t]=\operatorname{softmax}\left(\tilde{f}_{1}[t], \ldots, \tilde{f}_{|A|}[t]\right),
\end{equation}
where $\phi(\cdot)$ is a normalization (e.g., z-score or temperature scaling) and $\alpha \in[0,1]$ is a learnable or fixed mixing coefficient. The fused representation is then,
\begin{equation}
\mathbf{Z}[t]=\mathbf{F F N}\left(\sum_{s=1}^{|A|} w_{s}[t] \mathbf{v}_{s}[t]\right) \in \mathbb{R}^{Q},
\end{equation}
where $\operatorname{FFN}(\cdot)$ denotes a position-wise feed-forward network (FFN) that refines the fused representation. Then, $\operatorname{FFN}(\cdot)$ is fed into a prediction network $f_{\Theta}(\cdot)$ to produce beam probabilities $\hat{\mathbb{P}}=f_{\Theta}(\mathbf{Z}) \in \mathbb{R}^{Q}$, where $\Theta$ is model parameter set, and the final beam is $\hat{q}[t]=\arg \max _{q} \hat{\mathbf{P}}[q]$.

\emph{\textbf{End-to-end Learning Phase:}}
We design an end-to-end learning process. The unified training objective integrates task prediction accuracy and multi-modal alignment consistency, leading to a robust and generalizable framework. Specifically, we optimize beam index prediction accuracy through cross-entropy loss defined as:
\begin{equation}
\mathcal{L}_{1}=-\sum_{q=1}^{Q} p_{q}[t] \log \left(\sigma_{\text{Softmax}}\left(\hat{p}_{q}[t]\right)\right),
\end{equation}
where $p_{q} \in\{0,1\}$ denotes the ground-truth one-hot encoded optimal beam index, $\hat{p}_{q}$ represents the model prediction
probability for the $q$-th beam, and $\sigma_{\text{Softmax}}(\cdot)$ is the softmax
activation function. 

Moreover, the cross-modal feature similarity is achieved by normalizing the temperature-scaled contrast loss, i.e.,
\begin{equation}
\mathcal{L}_{2}=-\frac{\sum_{\gamma_{1}, \gamma_{2} \in A} p_{q}[t] \log \left(\sigma_{\text {Softmax }}\left(\frac{S_{\gamma_{1}, \gamma_{2}}[t]}{\theta}\right)\right)}{|A|(|A|-1)},
\end{equation}
where the temperature parameter $\theta>0$ regulates the similarity concentration: the smaller $\theta$, the sharper distribution; and the larger $\theta$, the softer distribution.

Thus, we design a trade-off objective subjected to by dynamic constraints on the contributions of different modal, which ensures beam prediction performance, as:
\begin{equation}
\mathcal{L}=\mathcal{L}_{1}+\beta \mathcal{L}_{2},
\end{equation}
where $\beta\ge0$ is trade-off parameter, controlling the contribution of different objectives.

\section{Experiments and Discussions} 


\subsection{Experimental Settings}
\emph{1) Dataset and Neural Network Architecture:} To evaluate performance, we utilize Scenario 23 in the DeepSense 6G dataset \cite{DeepSense}, which targets UAV communications and provides multi-modal information (RGB images, GPS, IMU, flight posture, velocity, and mmWave signals). The setup includes a fixed receiver with a 60 GHz phased-array antenna and camera, and a UAV with a mmWave transmitter. The original 64-dimensional beam power vectors are down-sampled to 32, with the maximum-power beam chosen as the ground-truth label. The dataset is divided into 70\% for training, 30\% for testing process. 
Based on this dataset, we design the model backbone using MLP architecture and ResNet blocks. 


\emph{2) Compared Methods:} We evaluate $\text{SaM²B}$ against three state-of-the-art baselines, described as follows: Based on NLinear \cite{zeng2022}, this method utilizes a linear model to map data. We incorporate it into our established multi-modal framework for modal linear fusion. To distinguish it from the original algorithm, we name it MM-NLinear. Based on BBOX scheme \cite{zheng2025}, this method employs full-image input to encode key visual data. We integrate this model into an existing modal linear fusion multi-modal framework while retaining our contrastive learning scheme. To distinguish it from the original algorithm, we name it No-BBOX. Moreover, we compared a scheme \cite{Charan2022} that utilizes image and location information to aid beam prediction, named as MM-aid.

To ensure fair comparisons, we standardized the experimental environments for all methods. Specifically, all models share the same neural network backbone architecture and are subject to equivalent constraints on computational complexity and memory usage to simulate resource-constrained edge device deployment scenarios.

\begin{figure}[t]
\centering
\includegraphics[width=1\linewidth]{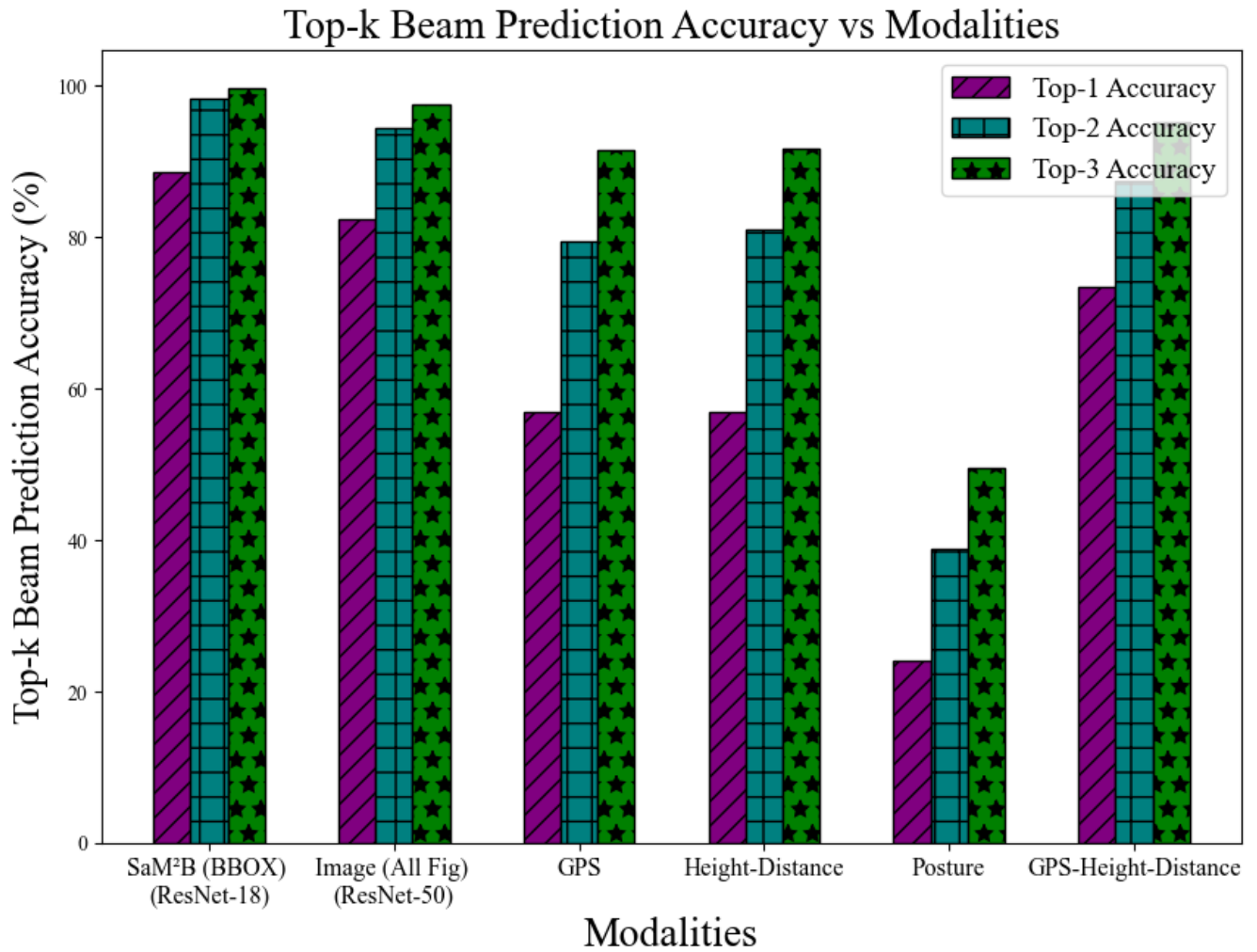}
\caption{Performance comparison under different modalities of the proposed $\text{SaM²B}$.}
\label{fig2}
\end{figure}

\begin{figure}[t]
\centering
\includegraphics[width=1\linewidth]{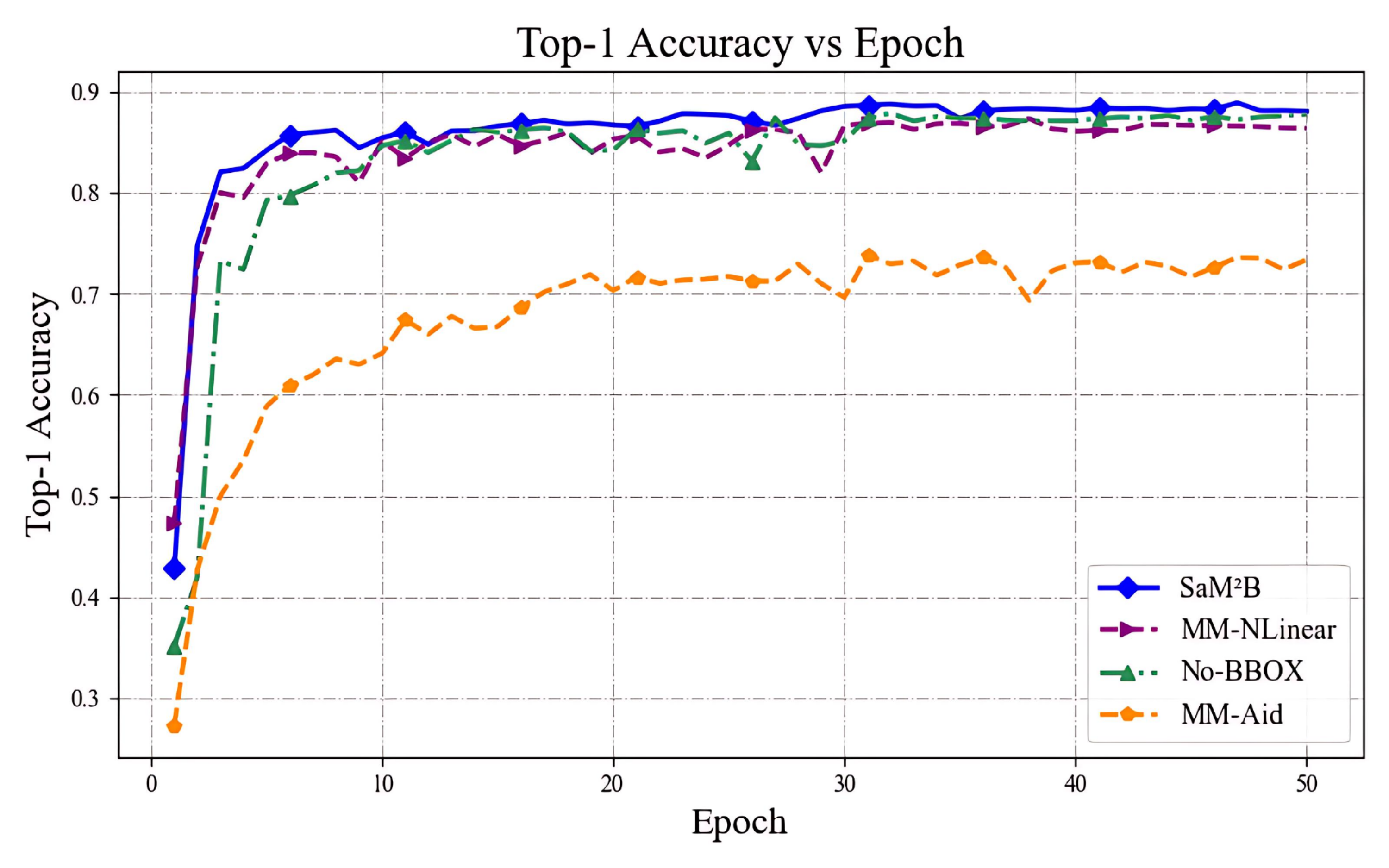}
\caption{Task performance comparison about $\text{SaM²B}$ and other baselines.}
\label{fig3}
\end{figure}

\subsection{Experimental Results on Task Inference Performance}
To evaluate the utility of different modalities in mmWave beam prediction, we conducted ablation experiments as shown Figure \ref{fig2}. Specifically, $\text{SaM²B}$ (using only the target BBOX, ResNet-18) achieved Top-1 = 88.63\%, Top-2 = 98.26\%, Top-3 = 99.68\%, slightly outperforming the Whole image approach (using the deeper ResNet-50 architecture) at 86.32\%, 96.34\%, 99.41\%. This indicates that preserving the local appearance of the target after background removal is more beneficial for identifying the optimal beam. Performance significantly degrades when relying solely on geometric/positional information: GPS achieves 56.97\%, 79.34\%, 91.38\%, Height-Distance achieves 56.81\%, 80.94\%, 91.67\%, while using posture alone yields only 24.06\%, 38.89\%, 49.43\%. This demonstrates that single geometric modality information imposes weak constraints on beam indexing and is susceptible to noise. Fusing geometric information (GPS-Height-Distance) significantly improves performance to 73.42\%, 87.36\%, 95.18\%, validating that complementarity among multi-source geometric signals enhances beam prediction on high-speed moving UAVs. In summary, incorporating contrastive learning alignment loss into the multi-modal framework enhances discriminative power, and enables efficient processing using lighter-weight networks.

As shown in Figure \ref{fig3}, $\text{SaM²B}$ model demonstrates outstanding convergence characteristics, achieving a peak Top-1 accuracy of nearly 90\%, significantly outperforming all baseline models. The key gain comes from our reliability-aware dynamic weighting, which adaptively up-weights informative modalities. The MM-NLinear method, which employs linear projection for multi-modal fusion, fails to account for the overlapping effects between different modalities, resulting in performance inferior to $\text{SaM²B}$. In contrast, the MM-Aid approach, which utilizes positional information, performs the worst, as geometric information alone does not yield improved beam prediction. The No-BBOX baseline, which relies solely on full-image encoding within the multi-modal contrastive learning framework, maintains high accuracy but suffers from high computational complexity due to processing the entire image. This highlights the critical role of structured spatial information for effective beam prediction.

\section{Conclusions}
In this paper, we propose a reliability-aware dynamic weighting mechanism for the semantic-aware multi-modal beamforming prediction framework ($\text{SaM²B}$). By integrating multi-modal environmental data and adaptively allocating the contributions of different modalities over time through reliability-aware weight updates, our method generates modality-invariant semantic representations to address the issues of modal mismatch and weak alignment. Experimental results on real-world low-altitude UAV datasets demonstrate that $\text{SaM²B}$ significantly outperforms traditional benchmark methods in communication performance. Future work include extensions through time prediction for active beam tracking, online uncertainty awareness to further evaluate generalization ability and robustness.

\bibliographystyle{IEEEtran} 
\bibliography{bib}

\end{document}